\documentclass[letterpaper]{article} 
\usepackage{aaai25}  
\usepackage{times}  
\usepackage{helvet}  
\usepackage{courier}  
\usepackage[hyphens]{url}  
\usepackage{graphicx} 
\usepackage{booktabs}
\usepackage{amssymb}
\usepackage{subfigure}
\usepackage{amsmath}
\newcommand{\correspondingauthor}{\textsuperscript{*}}
\usepackage{natbib}  
\usepackage{caption} 
\usepackage{algorithm}
\usepackage{algorithmic}

\usepackage{newfloat}
\usepackage{listings}
\DeclareCaptionStyle{ruled}{labelfont=normalfont,labelsep=colon,strut=off} 
\floatstyle{ruled}
\newfloat{listing}{tb}{lst}{}
\floatname{listing}{Listing}
\title{Doubly Stochastic Adaptive Neighbors Clustering \\via the Marcus Mapping}
\author{
    Jinghui Yuan\textsuperscript{\rm 1,\rm 2}, Chusheng Zeng\textsuperscript{\rm 1}, Fangyuan Xie\textsuperscript{\rm 1,\rm 2}, Zhe Cao\textsuperscript{\rm 1} \\ Mulin Chen\textsuperscript{\rm 1}, Rong Wang\textsuperscript{\rm 1}\thanks{Corresponding author}, Feiping Nie\textsuperscript{\rm 1,\rm 2}\correspondingauthor, Yuan Yuan\textsuperscript{\rm 1}
}
\affiliations{
    \textsuperscript{\rm 1}School of Artificial Intelligence, OPtics and ElectroNics (iOPEN),\\
 Northwestern Polytechnical University, Xi’an 710072, Shaanxi, China\\
    \textsuperscript{\rm 2}Institute of Artificial Intelligence (TeleAI), China Telecom, P. R. China \\
    yuanjh@mail.nwpu.edu.cn, zcs@mail.nwpu.edu.cn, xiefangyuan@mail.nwpu.edu.cn, caozhe@mail.nwpu.edu.cn\\ chenmulin001@gmail.com, wangrong07@tsinghua.org.cn, feipingnie@gmail.com, y.yuan1.ieee@gmail.com.

}

\usepackage{bibentry}

\begin{document}

\maketitle

\begin{abstract}
Clustering is a fundamental task in machine learning and data science, and similarity graph-based clustering is an important approach within this domain. Doubly stochastic symmetric similarity graphs provide numerous benefits for clustering problems and downstream tasks, yet learning such graphs remains a significant challenge. Marcus theorem states that a strictly positive symmetric matrix can be transformed into a doubly stochastic symmetric matrix by diagonal matrices. However, in clustering, learning sparse matrices is crucial for computational efficiency. We extend Marcus theorem by proposing the Marcus mapping, which indicates that certain sparse matrices can also be transformed into doubly stochastic symmetric matrices via diagonal matrices. Additionally, we introduce rank constraints into the clustering problem and propose the Doubly Stochastic Adaptive Neighbors Clustering algorithm based on the Marcus Mapping (ANCMM). This ensures that the learned graph naturally divides into the desired number of clusters. We validate the effectiveness of our algorithm through extensive comparisons with state-of-the-art algorithms. Finally, we explore the relationship between the Marcus mapping and optimal transport. We prove that the Marcus mapping solves a specific type of optimal transport problem.
\end{abstract}

%

\section{Introduction}
Clustering data with complex structure is an essential problem in data science, the goal of which is to divide the given samples into different categories. Graph-based clustering has been attracting increasing attention due to its ability to capture the intrinsic relationships between data points, allowing for more accurate and meaningful clustering results \cite{ju2023glcc}. It has been also extensively applied in classification, segmentation, protein sequences analysis and so on \cite{huang2023style, barrio2023clustering}.

Spectral clustering (SC) is a typical graph-based clustering method whose core idea is to construct a similarity graph and partition the samples by cutting this graph. Over the past decades, there have been many significant works on spectral clustering \cite{Nie2011,bai2020sparse,Zong2024Self}. According to different normalization approaches, SC could be divided into ratio cut (Rcut) and normalized cut (Ncut). However, if the similarity matrix is doubly stochastic, the results of Rcut and Ncut are actually the same. The doubly stochastic affinity graph could greatly benefit clustering performance, which could also contribute to dimensionality reduction \cite{van2024snekhorn}, transformers \cite{sander2022sinkformers} and reaction prediction \cite{meng2023doubly}. 

Meanwhile, \cite{zass2006doubly} highlights the positive impact of doubly stochastic matrices on clustering results and subsequent tasks. \cite{pmlr-v162-ding22a} establishes conditions under which projecting an affinity matrix onto doubly stochastic matrices improves clustering by ensuring ideal cluster separation and connectivity. Consequently, extending clustering algorithms to incorporate doubly stochastic matrices has become a hot topic. The key issue lies in how to obtain a doubly stochastic matrix. Some researchers have proposed several methods to address this issue.

For instance, \cite{zass2005unifying} propose a method to transform a non-negative matrix into a doubly stochastic matrix. They believe that by iterating the process $K^{(t+1)}\leftarrow D^{-1/2}K^tD^{-1/2}$ with $D=diag(K^{t}1)$, the sequence converges to a doubly stochastic matrix. However, this method suffers from slow convergence, numerical instability, sensitivity to initialization, and high computational complexity for large matrices. \cite{zass2006doubly} present a Frobenius-optimal doubly stochastic normalization method via Von-Neumann successive projection and apply it into SC. \cite{marcus1961permanent} proposed a similar theorem, stating that any positive symmetric matrix can be transformed by multiplying it on the left and right with the same diagonal matrix. Therefore, it is valuable to explore and extend Marcus's theory, providing good conditions and algorithms for this transformation.

Optimal transport theory has gained significant attention in recent years due to its wide applicability in various fields. \cite{villani2003topics} laid the foundational mathematical framework for optimal transport, focusing on the cost of transporting mass in a way that minimizes the overall transportation cost. \cite{cuturi2013sinkhorn} revolutionized the field by introducing the Sinkhorn algorithm \cite{sinkhorn1967concerning}, which employs entropy regularization to make the computation of optimal transport more efficient and scalable to high-dimensional data. \cite{peyre2019computational} further developed algorithms for computational optimal transport, enhancing its practicality for large-scale problems. However, the relationship between clustering and optimal transport has been relatively unexplored \cite{yan2024optimal}. Introducing the concept of optimal transport into clustering is valuable and meaningful.

\cite{nie2016constrained} propose the rank constraint, which states that if the Laplacian matrix of the learned similarity matrix has a rank of $n-c$, then the similarity matrix has exactly $c$ connected components. The rank constraint clustering has also been applied in \cite{nie2017self,WANG2022108517}. However, the learned matrix is generally not a doubly stochastic symmetric matrix, which means it differs from a probability matrix. In this paper, we propose the Doubly Stochastic Adaptive Neighbors Clustering algorithm based on the Marcus Mapping (ANCMM). Our method can learn symmetric doubly stochastic similarity matrix, also known as probability matrices, which naturally have exactly $c$ connected components, allowing direct determination of clustering results without post-processing. Besides, we extend Marcus theorem by introducing a more relaxed constraint, proposing the Marcus mapping and proving that it can transform certain sparse non-negative matrices into symmetric doubly stochastic matrices through diagonal matrices. Additionally, we explore the relationship between the Marcus mapping and optimal transport, proving that the Marcus mapping solves a specific optimal transport problem. We summarize our main contributions below:
\begin{itemize}
	\item We propose the Marcus mapping theorem, the conditions for which have been rigorously proven. An iterative method to compute the Marcus mapping theorem is also provided. This extends the Marcus theorem by relaxing its requirement for positive matrices.
	\item We explore the relationship between the Marcus mapping algorithm and optimal transport, proving that the Marcus mapping algorithm is a special case of optimal transport. 
	\item We propose the Doubly Stochastic Adaptive Neighbors Clustering method based on the Marcus Mapping (ANCMM) and design an optimization algorithm to solve this problem. The proposed method empirically converges on the tested datasets. We validate the effectiveness of our method through extensive comparative experiments on synthetic and real-world datasets.
\end{itemize}

\section{Methodology}
In this section, we first introduce the notations used in the paper, then proceed to describe and prove the generalized Marcus mapping and adaptive neighbors. Finally, we present the optimization problem to be solved.

\subsection{Notations}
Throughout this paper, data matrix is denoted as $X\in\mathbb{R}^{n \times d}$, where $n$ is the number of samples. $d$ is the number of features. $x_i\in\mathbb{R}^d$ is the $i$-th sample and also the transpose of the $i$-th
row of $X$. $\mathbb{I}_n$ or $\mathbb{I}$ denotes a vector in the space $\mathbb{R}^n$ with all the elements being 1. $\pi$ represents a full permutation. For example, given a permutation rule $\pi$, $\{\pi(1),\pi(2),...,\pi(n)\}$ is just $\{1,2,...,n\}$ but $\pi(1),\pi(2),...,\pi(n)$ and $1,2,...,n$ are in a different order. For matrix $S$, $Tr(S)$ represents the trace of 
$S$, and $\|S\|_F$ represents the Frobenius norm of $S$ .

\subsection{Marcus Mapping}

In this section, we introduce the Marcus mapping. The Marcus mapping is a generalization of the Marcus theorem from strictly positive matrices to sparse non-negative matrices satisfying a weaker condition.

\subsubsection{Marcus theorem}\cite{marcus1961permanent}
If \(S\in\mathbb{R}^{n\times n}\) is symmetric and all its entries are strictly positive, then there exists a diagonal matrix \(D\) with positive main diagonal entries such that \(DSD\) is doubly stochastic.

In other words, for a symmetric matrix \(S\) satisfying the strict positivity condition, there exists a diagonal matrix \(D\) such that
\(
\mathcal{M}(S)=DSD
\) and
\begin{equation}
\mathcal{M}(S)\in
\{Z\in\mathbb{R}^{n\times n}\mid
Z\mathbb{I}_n=\mathbb{I}_n,\ Z_{ij}\geq0,\ Z=Z^T\}.
\end{equation}

In this paper, we extend the Marcus theorem by relaxing the requirement for strictly positive matrices to symmetric non-negative matrices.

\cite{zass2005unifying} proposed a similar idea. They considered transforming a non-negative symmetric matrix \(S\) into a doubly stochastic matrix by iteratively multiplying \(S\) on the left and right by a degree matrix \(D^{-\frac{1}{2}}\), where

\begin{equation}
D=diag(S\mathbb{I}_n).
\end{equation}

However, this transformation cannot be guaranteed for an arbitrary non-negative symmetric matrix. For example, consider the following matrix:

\begin{equation}
S=
\begin{pmatrix}
0&0&1\\
0&0&1\\
1&1&0
\end{pmatrix}.
\end{equation}

For any diagonal matrix \(D\) with positive main diagonal entries, \(DSD\) has the same zero pattern as \(S\). The matrix \(S\) does not have total support, and therefore \(DSD\) cannot be doubly stochastic for any choice of \(D\).

To characterize the condition required for the Marcus mapping, we first introduce the definition of \(\pi\)-diagonals and the Sinkhorn theorem.

\subsubsection{Definition 1 (\(\pi\)-diagonals)}

If \(S\in\mathbb{R}^{n\times n}\) and \(\pi\) is a permutation of \(\{1,2,\ldots,n\}\), then the sequence

\begin{equation}
(S_{1,\pi(1)},S_{2,\pi(2)},\ldots,S_{n,\pi(n)})
\end{equation}

is called the \(\pi\)-diagonal of \(S\) corresponding to \(\pi\). The matrix \(S\) is said to have total support if \(S\ne0\) and every positive element of \(S\) lies on a positive \(\pi\)-diagonal. A non-negative matrix that contains a positive \(\pi\)-diagonal is said to have support.

If \(\pi\) is the identity permutation, the corresponding \(\pi\)-diagonal is called the main diagonal.

\subsubsection{Theorem 1 (Sinkhorn theorem)}

If \(S\in\mathbb{R}^{n\times n}\) is non-negative, there exist diagonal matrices \(D_1\) and \(D_2\) with positive main diagonal entries such that \(\hat{S}=D_1SD_2\) is doubly stochastic if and only if \(S\) has total support. If \(\hat{S}\) exists, then it is unique \cite{sinkhorn1967concerning}.

\subsubsection{Theorem 2 (Marcus mapping theorem)}
Let \(S\in\mathbb{R}^{n\times n}\) be a symmetric non-negative matrix. If
\(
S_{ii}>0,\quad i=1,\ldots,n,
\) then there exists a diagonal matrix \(D\) with positive main diagonal entries such that
\(
\mathcal{M}(S)=DSD\in
\{Z\in\mathbb{R}^{n\times n}\mid
Z\mathbb{I}_n=\mathbb{I}_n,\ Z_{ij}\geq0,\ Z=Z^T\}.
\)

\subsubsection{Proof}

We first show that \(S\) has total support. For every positive diagonal element \(S_{ii}>0\), the element \(S_{ii}\) lies on the positive \(\pi\)-diagonal corresponding to the identity permutation. For any positive off-diagonal element \(S_{ij}>0\), where \(i\ne j\), consider the permutation:
\begin{equation}
\pi(i)=j,\qquad
\pi(j)=i,\qquad
\pi(k)=k,\quad k\notin\{i,j\}.
\end{equation}
Since \(S\) is symmetric, \(S_{ji}=S_{ij}>0\). Moreover, \(S_{kk}>0\) for every \(k\). Therefore, the \(\pi\)-diagonal corresponding to this permutation is positive. Hence, every positive element of \(S\) lies on a positive \(\pi\)-diagonal, and \(S\) has total support.

According to Theorem 1, there exist diagonal matrices \(D_1\) and \(D_2\) with positive main diagonal entries such that \(D_1SD_2\) is doubly stochastic. Since \(S=S^T\), 
\((D_1SD_2)^T=D_2SD_1\) is also doubly stochastic. By the uniqueness of the doubly stochastic matrix in Theorem 1,
\(
D_1SD_2=D_2SD_1.
\)

Let \(D\) be the diagonal matrix whose \(i\)-th diagonal entry is
\(
D_{ii}=\sqrt{(D_1)_{ii}(D_2)_{ii}}.
\)

The equality above implies that \(DSD=D_1SD_2\) on the support of \(S\). Outside the support of \(S\), both matrices have zero entries. Therefore,
\(
DSD=D_1SD_2.
\)

Thus, \(DSD\) is doubly stochastic. Since \(S\) and \(D\) are symmetric,
\(
(DSD)^T=DSD.
\)
Therefore, \(\mathcal{M}(S)=DSD\) is a symmetric doubly stochastic matrix. \(\Box\)

The computational procedure of the Marcus mapping is summarized in Algorithm 1. The algorithm alternately updates \(D_1\) and \(D_2\) until \(D_1SD_2\) becomes doubly stochastic, and then constructs the symmetric mapping \(\mathcal{M}(S)=DSD\).

\begin{algorithm}
    \caption{Marcus mapping algorithm}
    \label{Algorithm 1}
    \textbf{Input}: Symmetric non-negative matrix \(S\in\mathbb{R}^{n\times n}\) \\
    \textbf{Output}: \(\mathcal{M}(S)\)
    \begin{algorithmic}[1]
        \STATE Initialize \(D_1=D_2=diag(\mathbb{I}_n)\)
        \WHILE{not converge}
            \STATE \(D_1=diag\left(1./(SD_2\mathbb{I}_n)\right)\)
            \STATE \(D_2=diag\left(1./(S^TD_1\mathbb{I}_n)\right)\)
        \ENDWHILE
        \STATE \(D_{ii}=\sqrt{(D_1)_{ii}(D_2)_{ii}},\quad i=1,\ldots,n\)
        \STATE \(\mathcal{M}(S)=DSD\)
    \end{algorithmic}
\end{algorithm}

\subsubsection{Corollary 1}
Let $S\in\mathbb{R}^{n\times n}$ be a symmetric non-negative matrix.
If $S$ has total support, then there exists a diagonal matrix $D$ with positive diagonal entries such that
\(
\mathcal{M}(S)=DSD
\)
is a symmetric doubly stochastic matrix.

\subsubsection{Corollary 2}
Let $S\in\mathbb{R}^{n\times n}$ be a symmetric non-negative matrix with total support. Consider the iterative scheme
\(
u^{(t+1)}=\mathbf{1}./(Su^{(t)}),
\)
where $u^{(0)}>0$. Suppose that the sequence converges to a positive vector $u^*$ satisfying
\(
u^*=\mathbf{1}./(Su^*).
\)
Define
\(
D=\operatorname{diag}(u^*),\ 
\widetilde S=DSD.
\)
Then $\widetilde S$ is a symmetric doubly stochastic matrix. Moreover,
\(
\widetilde S=\mathcal{M}(S).
\)

The proofs of Corollaries 1 and 2 are both relatively straightforward. Corollary 1 follows directly from the Sinkhorn theorem and the construction of an identical diagonal scaling matrix in the Marcus mapping. For Corollary 2, once the iterative scheme converges, the resulting matrix is identical to that obtained by the Marcus mapping.

\subsection{Adaptive Local  doubly stochastic Structure Learning}
In graph-based learning, an important aspect is to find a low-dimensional local manifold representation, as high-dimensional data generally contain low-dimensional manifolds. Therefore, selecting an appropriate similarity graph is a crucial requirement for graph-based learning. A reasonable objective function for graph learning is shown below \cite{nie2014clustering}.
\begin{equation}
\begin{aligned}
\label{eq8}
&\min _{s_{i} \in \mathbb{R}^{n \times 1}} \sum_{i, j}^{n}\left\|x_{i}-x_{j}\right\|_{2}^{2} s_{i j}+\alpha\|S\|_{F}^{2} \\
&\text { s.t. } \quad \forall i, s_{i}^{T} \mathbb{I}=1,0 \leq s_{i j} \leq 1
\end{aligned}
\end{equation}

In the research process, $s_{ij}$ is usually considered the probability that $x_i$ and $x_j$ belong to the same class. However, the solution to this problem is not a doubly stochastic matrix, which cannot reasonably express the concept of probability. Therefore, we impose a stronger constraint on problem \eqref{eq8}
\begin{equation}
\begin{aligned}
\label{eq9}
&\min _{s_{i} \in \mathbb{R}^{n \times 1}} \sum_{i, j}^{n}\left\|x_{i}-x_{j}\right\|_{2}^{2} s_{i j}+\alpha\|S\|_{F}^{2} \\
&\text { s.t. } \quad S^{T} \mathbb{I}=\mathbb{I},0 \leq s_{i j} \leq 1,S=S^{T}
\end{aligned}
\end{equation}
Typically, the learned probability matrix $S$ does not have $c$ connected components, requiring further clustering using methods like K-means. Due to the following theorem, we introduce the Laplacian rank constraint.
\subsubsection{Theorem 3.} For a nonnegative symmetric matrix \(S\), if the rank of the Laplacian matrix of \(S\) is $n-c$, i.e. $rank(L_S)=n-c$, then $S$ has exactly $c$ connected components.

Hence, the Laplacian rank constraint is introduced, and the final form of the loss function becomes:
\begin{equation}
\begin{aligned}
\label{eq10}
&\min _{s_{i} \in \mathbb{R}^{n \times 1}} \sum_{i, j}^{n}\left\|x_{i}-x_{j}\right\|_{2}^{2} s_{i j}+\alpha\|S\|_{F}^{2} \\
&\text { s.t. }\ S^{T} \mathbb{I}=\mathbb{I},0 \leq s_{i j} \leq 1,S=S^{T},rank(L_S)=n-c
\end{aligned}
\end{equation}

Problem \eqref{eq10} is our objective function. We have designed an optimization algorithm to solve this problem.

\section{Optimization Algorithm}
Optimizing problem \eqref{eq10} is more challenging because the matrix constraints are coupled. However, we can cleverly solve this problem by iteratively solving the relaxed problem and using the Marcus mapping.
\subsection{Clustering}

Assume $\sigma_i(L_S)$ represents the 
i-th smallest eigenvalue of  $L_S$. Since 
$L_S$ is a positive semidefinite matrix, 
$\sigma_i(L_S)\ge 0$ . Because solving for the constraint $rank(L_s)=n-c$ is too difficult, we relax it to $\sum^c_{i=1}\sigma_i(L_S)=0$, which means $\sigma_i(L_S)=0 \ \forall \ i=1...c$.
According to Ky Fan’s Theorem \cite{Fan1949OnAT}, we have
\begin{equation}
\begin{aligned}
\sum_{i=1}^{c} \sigma_{i}\left(L_{S}\right)=\min _{F \in \mathbb{R}^{n \times c}, F^{T} F=I} Tr\left(F^{T} L_{S} F\right)
\end{aligned}
\end{equation}

Then problem (\ref{eq10}) is equivalent to the following problem
\begin{equation}
\begin{aligned}
&\min _{s_{i} \in \mathbb{R}^{n \times 1}} \sum_{i, j}^{n}\left\|x_{i}-x_{j}\right\|_{2}^{2} s_{i j}+\alpha\|S\|_{F}^{2}+2\lambda Tr(F^TL_SF)\\
&\text { s.t. } \quad S^{T} \mathbb{I}=\mathbb{I},0 \leq s_{i j} \leq 1,S=S^{T},F^TF=I
\end{aligned}
\end{equation}

We selected a sufficiently large $\lambda$, so that in the optimization process, $\sum^c_{i=1}\sigma_i(L_S)=0$ can be obtained, thereby solving the relaxed constraint.

\subsubsection{Fix $S$, update $F$}
Once $S$ is fixed, updating $F$ only requires solving the following problem
\begin{equation}
\begin{aligned}
\label{eq13}
\min _{F \in \mathbb{R}^{n \times c}, F^{T} F=I} Tr\left(F^{T} L_{S} F\right)
\end{aligned}
\end{equation}
The optimal solution for $F$ consists of the eigenvectors corresponding to the smallest 
c eigenvalues of $L_S$. This is part of Ky Fan's theorem.
\subsubsection{Fix $F$, update $S$}
Due to the following important equation
\begin{equation}
\sum_{i,j}\|f_i-f_j\|_2^2s_{ij}=2Tr(F^TL_SF)
\end{equation}
we are essentially required to solve the following problem:
\begin{equation}
\begin{aligned}
\label{eq15}
&\min _{s_{i} \in \mathbb{R}^{n \times 1}} \sum_{i, j}^{n}(\left\|x_{i}-x_{j}\right\|_{2}^{2}+\lambda\left\|f_{i}-f_{j}\right\|_{2}^{2}) s_{i j}+\alpha\|S\|_{F}^{2}\\
&\text { s.t. } \quad  S^{T} \mathbb{I}=\mathbb{I},0 \leq s_{i j} \leq 1,S=S^{T}
\end{aligned}
\end{equation}

Solving this problem is very challenging. \textbf{Our key idea is to first solve the relaxed problem and then map it to a symmetric doubly stochastic matrix using the Marcus mapping.} In this section, we first consider the relaxed problem.
\begin{equation}
\begin{aligned}
\label{Eq16}
&\min _{s_{i} \in \mathbb{R}^{n \times 1}} \sum_{i, j}^{n}(\left\|x_{i}-x_{j}\right\|_{2}^{2}+\lambda\left\|f_{i}-f_{j}\right\|_{2}^{2}) s_{i j}+\alpha\|S\|_{F}^{2}\\
&\text { s.t. } \quad  s_{i}^{T} \mathbb{I}=1,0 \leq s_{i j} \leq 1
\end{aligned}
\end{equation}

Denote $m_{ij}=\|x_i-x_j\|_2^2+\lambda\|f_i-f_j\|_2^2$ and let $m_i \in \mathbb{R}^{n \times 1}$ with $(m_i)_j=m_{ij}$. Because we have relaxed the constraints and decoupled the row‑column relationships, each row can be solved independently. By expanding and simplifying the original problem, we finally reduce it to the following form with a closed‑form solution.
\begin{equation}
\begin{aligned}
\label{eq17}
\min _{s_{i}} \quad\left\|s_{i}+\frac{1}{2 \alpha} m_{i}\right\|_{2}^{2} \quad \text { s.t. } \quad s_{i}^{T} \mathbb{I}=1,0 \leq s_{i j} \leq 1
\end{aligned}
\end{equation}
By solving this problem, we obtain the optimal solution 
$S^*_r$ after relaxation. We transform it into a symmetric matrix using the following symmetrization to meet the conditions for the Marcus mapping. 
\begin{equation}
\begin{aligned}
\label{eq18}
\hat{S^*_r}\leftarrow \frac{S^*_r+S^{*,T}_r}{2}
\end{aligned}
\end{equation}

We will later provide a heuristic justification: \textbf{1. Prove that it is easy to satisfy the conditions of the Marcus mapping by choosing $\alpha$. 2. Prove that the solution obtained through Marcus mapping is sufficiently close to the true solution of problem \eqref{eq15}.}

\subsubsection{Fix $F$, update $\mathcal{M}(S)$}
At this step, we utilize the Marcus mapping to compute $\mathcal{M}(S)$. Specifically, we input $\hat{S}^*_r$ into the Marcus mapping and aim to obtain
\begin{equation}
    S\leftarrow\mathcal{M}(\hat{S}^*_r)=D\hat{S}^*_rD=D\frac{S^*_r+S^{*,T}_r}{2}D
\end{equation}

After calculating $S=\mathcal{M}(\hat{S}^*_r)$, we substitute it into the first step and iterate through various steps until convergence. We summarize our optimization algorithm in Algorithm 2.

\begin{algorithm}
	\caption{ANCMM}
	\label{Algorithm 2} 
        \textbf{Input}: Data matrix $X \in \mathbb{R}^{n\times d}$, clusters $c$\\
        \textbf{Parameter}: $\alpha$ and $\lambda_0$\\
        \textbf{Output}: Desired similarity graph $S \in \mathbb{R}^{n\times n}$ with $c$ connected components
	\begin{algorithmic}[1]
            \STATE Initialize $S$
            \WHILE{not converge}
    		\STATE Fix $S$ and update $F$ by optimizing problem \eqref{eq13}
            \STATE Fix $F$ and update $S$ by solving problem \eqref{Eq16}
            \STATE Symmetrize matrix $S$ by Eq. \eqref{eq18}            
            \STATE Update $S$ by Marcus mapping using Algorithm 1
	    \ENDWHILE
	\end{algorithmic}
\end{algorithm}

\subsection{Select $\alpha$ based on sparsity}
In this part, we will demonstrate how the selection of 
$\alpha$ affects the sparsity of the probability matrix $S$ . Additionally, we will prove that by simply choosing $\alpha$ we can satisfy the conditions required by the Marcus mapping.

We still consider the relaxed problem \ref{Eq16}. Because if  the relaxation problem's $\alpha$ can control the sparsity of the matrix $S$, the non-relaxed one can too. Since we update each row individually, \cite{nie2014clustering} has proven that the closed-form solution has the following form.
\begin{equation}
\label{Eq20}
s_{i j}=\left(-\frac{m_{i j}}{2 \alpha_{i}}+\phi\right)_{+}
\end{equation}

where $\phi=\frac{1}{k}+\frac{1}{2 k \alpha_{i}} \sum_{j=1}^{k} m_{i j} $. That  $x_{i}$  has  k  neighbours $\longleftrightarrow s_{i j}>0, \forall 1 \leqslant   j \le k $  and $ s_{i, k+1}=0$. Here, we assume $s_{ij}$ is sorted in descending order with respect to $j$. According to \eqref{Eq20} and the value of $\phi$ , we have
\begin{equation}
\frac{k}{2} m_{i k}-\frac{1}{2} \sum_{j=1}^{k} m_{i j}<\alpha_{i} \leq \frac{k}{2} m_{i, k+1}-\frac{1}{2} \sum_{j=1}^{k} m_{i j}
\end{equation}
where  $(m_{ij})=m_{i 1}, m_{i 2}, \cdots, m_{i n}$  are also sorted in ascending order with respect to $j$. 
This means that by controlling $\alpha_i$ for each row, we can precisely control the sparsity of each row.
\subsection{Heuristic Approximation Analysis}
In this section, we will demonstrate that $\mathcal{M}(\hat{S}_r^*)$ is sufficiently close to the true solution of problem \eqref{eq15}.

For problem \eqref{eq8}, due to the Karush-Kuhn-Tucker (KKT) conditions, there exist $\theta$ and $\gamma$ such that:
\begin{equation}
\label{eq23}
M+2\theta S^*_r+\gamma\mathbb{I}^T=0
\end{equation}
where $M_{ij}=m_{ij}$, $\theta\ge0$ and $\gamma \in \mathbb{R}^{n}$ , Similarly, considering the transpose problem of problem \eqref{eq8}, we have:
\begin{equation}
\label{eq24}
M+2\theta S^{*,T}_r+\mathbb{I}\gamma^T=0
\end{equation}

We know that 
\begin{equation}
\begin{aligned}
&\mathcal{M}(\frac{S^*_r+S^{*,T}_r}{2})\\=&D\frac{S^*_r+S^{*,T}_r}{2}D=\frac{1}{2} \left( DS^*_rD+DS^{*,T}_rD\right)
\end{aligned}
\end{equation} 

If $DS^*_rD,DS^{*,T}_rD$ are the optimal solutions to problem \eqref{eq8} and the transpose problem, respectively, then according to Eq. \eqref{eq23} and Eq. \eqref{eq24}, $\mathcal{M}(\hat{S}_r^*)$ is the optimal solution to problem \eqref{eq9}. However, this is usually not the case. Nevertheless, we can have the following estimation:
\begin{equation}
\begin{aligned}
&\|\mathcal{M}(\hat{S}_r^*)-\hat{S}_r^*\|_F^2\\&\le\frac{1}{2}(\|DS^*_rD-S^*_r\|_F^2+\|DS^{*,T}_rD-S^{*,T}_r\|_F^2)\\&=\|DS^*_rD-S^*_r\|_F^2\le\epsilon
\end{aligned}
\end{equation} 
Usually, $\epsilon$ is very small relative to $\|S^*_r\|_F^2$, making it an acceptable approximation. $\Box$

\subsection{Time Complexity Analysis}
When updating $F$ by Eq. \eqref{eq13}, c smallest eigenvalues need to be computed, Using the Lanczos algorithm can achieve a time complexity that is linear with respect to $\mathcal{O}(n)$. Then, the corresponding eigenvectors need to be computed, resulting in a time complexity of $\mathcal{O}(k_2n^2)$ , $k_2$ represents the number of iterations. 

When updating $S$ by the Eq. \eqref{eq17}, only $n$ linear equations need to be solved and a matrix addition needs to be calculated, resulting in a time complexity of $\mathcal{O}(n^2)$ . Then we calculate Marcus mapping by Algorithm 1, which only requires computing the multiplication of a matrix by a vector and the division of a scalar by a vector, resulting in a time complexity of $\mathcal{O}(k_1n^2)$. Here, $k_1$ represents the number of iterations. Therefore, the overall time complexity is $\mathcal{O}(k(k_1+k_2)n^2)$, $k$ represents the total number of iterations.

\section{Connection with Optimal Transport}
In this section, we will point out the connection between the Marcus mapping and optimal transport.

Denote the Marcus mapping for a non-negative matrix 
$S$ as $\mathcal{M}(S)$ , $\log(S)$ represents the natural logarithm of each element of $S$ , and we require $\log(0)=-\infty$ and $e^{-\infty}=0$.

Consider the optimal transport problem with entropy regularization:
\begin{equation}
\begin{aligned}
&\min_{P}<P,-\log(S)>+\sum_{i j}\frac{1}{\omega}P_{ij}\log(P_{ij})\\
&\text { s.t. } \quad P\in \Omega(\mathbb{I},\mathbb{I})
\end{aligned}
\end{equation} 
where $\Omega(\mathbb{I},\mathbb{I})=\{Z|Z_{ij}\ge0,Z\mathbb{I}=\mathbb{I},Z^T\mathbb{I}=\mathbb{I}\}$

The Lagrangian function of the optimal transport problem is given by
\begin{equation}
\begin{aligned}
\mathcal{L}(P, \phi, \xi)&=\sum_{i j} \left(\frac{1}{\omega} p_{i j} \log p_{i j}-p_{i j} \log(s_{ij})\right)\\&+\phi^{T}\left(P \mathbb{I}_{n}-\mathbb{I}_{n}\right)+\xi^{T}\left(P^{T} \mathbb{I}_{n}-\mathbb{I}_{n}\right)
\end{aligned}
\end{equation} 

By taking the partial derivative with respect to $P$, we can obtain that the optimal solution satisfies:
\begin{equation}
\begin{aligned}
p_{ij}(\omega)=e^{-\frac{1}{2}-\omega\phi_i}(s_{ij})^{\omega}e^{-\frac{1}{2}-\omega\xi_j}
\end{aligned}
\end{equation} 

In particular, when $\omega$ is 1, the form of the optimal transport solution is consistent with the Marcus mapping. 

\begin{equation}
\begin{aligned}
p_{ij}|_{\omega=1}=e^{-\frac{1}{2}-\phi_i}s_{ij}e^{-\frac{1}{2}-\xi_j}
\end{aligned}
\end{equation} 

Since $S$ is symmetric, and
\begin{equation}
\begin{aligned}
P=\underset{P\in \Omega(\mathbb{I},\mathbb{I})}{argmin}\left(<P,-\log(S)>+\sum_{ij}\frac{1}{\omega}P_{ij}\log(P_{ij})\right)
\end{aligned}
\end{equation} 
is also symmetric. Therefore, $P$ has the form $P=DSD$. This demonstrates that the Marcus mapping is solving a special optimal transport problem.

The transformation of $-\log(S)$ is crucial as it can convert zeros in the probability matrix to $+\infty$, thereby satisfying the conditions for optimal transport. \textbf{However, computing through the Marcus mapping is much more efficient than directly using optimal transport, primarily because optimal transport involves additional logarithmic computations, and expressing $+\infty$ is challenging.}

\section{Experiments on toy datasets}
To demonstrate the superiority of our algorithm, we first constructed a toy dataset in the shape of double moons. We set the sample size to 200, with a random seed of 1 and noise level of 0.13. We performed clustering using both the CAN and ANCMM algorithms, both of which incorporate rank constraints and are adaptive neighbors algorithms. We compared their accuracy and the configuration of adaptive neighbors. The dataset and results are shown in Figure 1.

\begin{figure}[h]
\label{F6}
        \centering
        \subfigure[Dataset]
	{
		\includegraphics[width=0.22\textwidth]{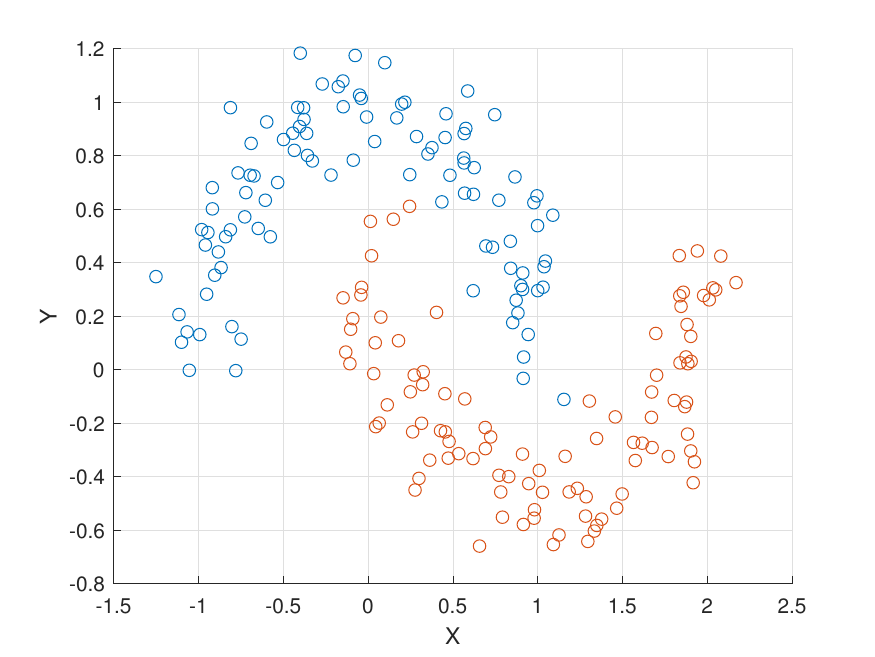}
	}
         \subfigure[Gaussian matrix]
	{
		\includegraphics[width=0.22\textwidth]{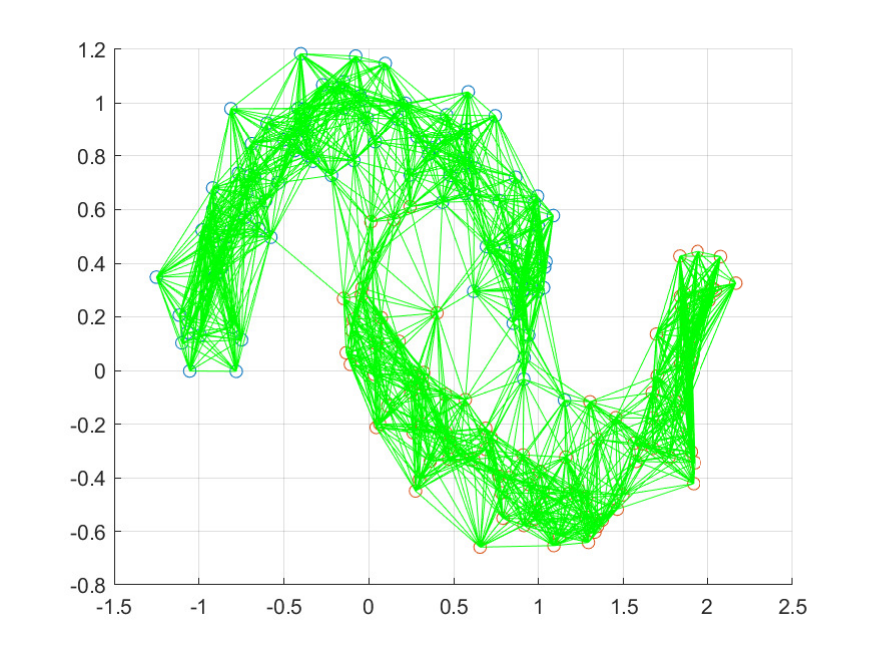}
	}
	\subfigure[CAN]
	{
		\includegraphics[width=0.22\textwidth]{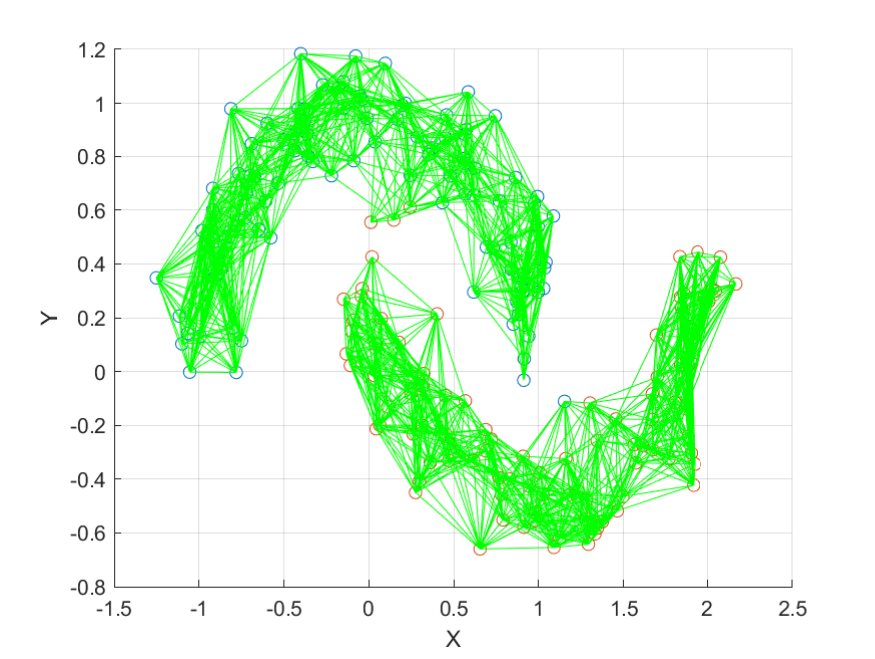}
	}
        \subfigure[CAN's error]
	{
		\includegraphics[width=0.22\textwidth]{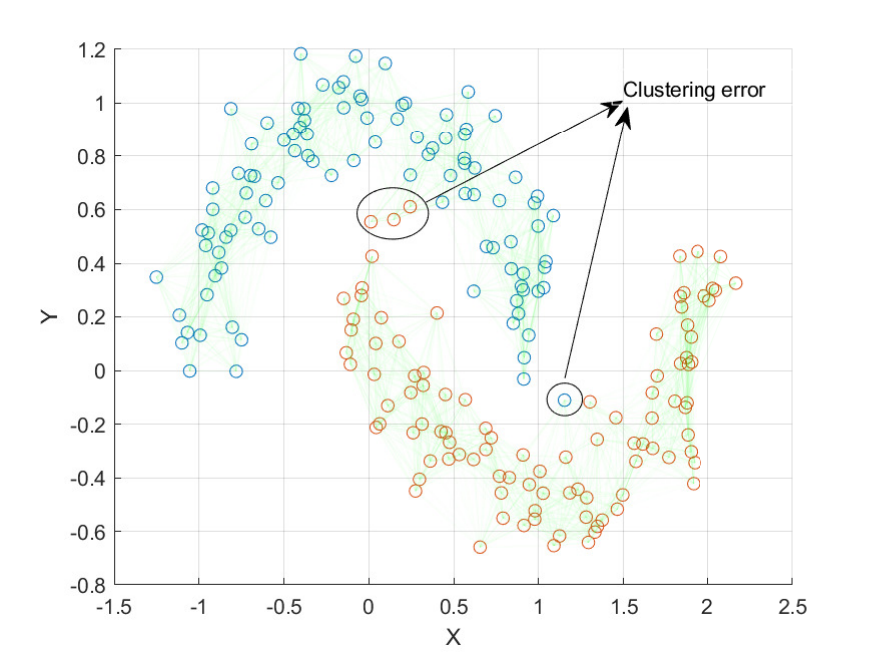}
	}
         \subfigure[ANCMM]
	{
		\includegraphics[width=0.22\textwidth]{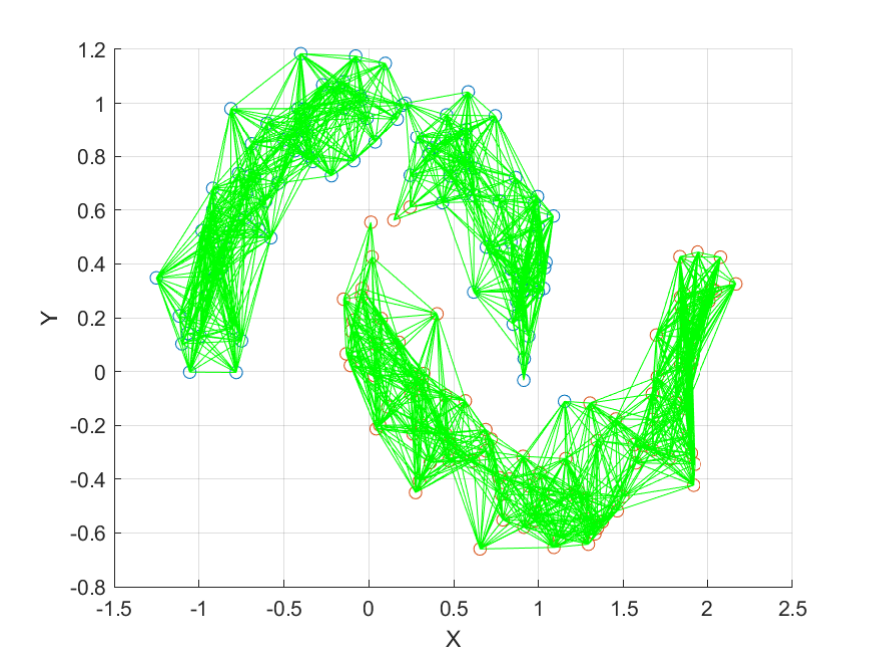}
	}
        \subfigure[ANCMM's error]
	{
		\includegraphics[width=0.22\textwidth]{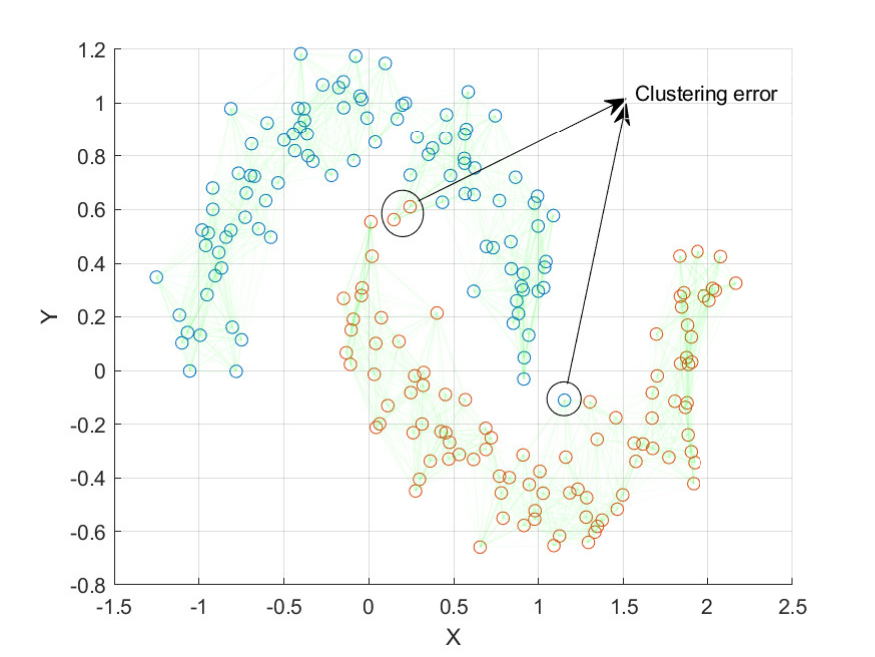}
	}
	\caption{Visualization on toy data. (a) Original data. (b) The Gaussian affinity matrix. (c) Result of CAN method. (d) CAN clustering error samples. (e) Result of ANCMM method. (f) ANCMM clustering error samples.}
\end{figure}

\begin{table}[!htbp]
	\centering
	\scriptsize
	\caption{Metrics on Toy Datasets }
	\label{T1}
	\scalebox{0.8}{
		\begin{tabular}{ccccccccc}
			\toprule
			Metric     & CAN &  ANCMM &Metric     & CAN &  ANCMM &Metric     & CAN &  ANCMM \\ 
			\midrule
			ACC	  & 98.0	& \textbf{98.5}&NMI	  & 86.2	& \textbf{88.9}		&
			PUR   & 98.0    & \textbf{98.5}    	\\		
			\bottomrule
	\end{tabular}}
\end{table}
We used the learned probability matrix weights as the edge weights connecting each pair of points. Additionally, we demonstrate the similarity matrix obtained solely from the Gaussian kernel function. Both CAN \cite{nie2014clustering} and ANCMM learn a similarity matrix with two connected components. However, as shown in the figure, CAN misclassified one more point compared to ANCMM. Table 1 shows the accuracy (ACC), normalized mutual information (NMI), and purity (PUR) metrics for the clustering results.

\section{Experiments on real datasets}
\subsection{Experimental Settings}
\subsubsection{Datasets.}
We conducted extensive experiments on ten datasets, including TR41, ORL, ARsmall, LetterRecognition, Yale, Wine, Feret, Movement, Ecoli, Yeast. Table \ref{T1} lists the number of samples and clusters for each dataset.

\begin{table}[!htbp]
	\centering
	\caption{Description of the benchmark datasets}
	\label{T1}
	\scalebox{0.8}{
		\begin{tabular}{@{}cccc@{}}
			\toprule
			Datasets     & $\#$Object & $\#$Attribute & $\#$Class \\ 
			\midrule
                TR41	  & 878		& 7454		& 10	\\
                ORL	  & 400		& 1024		& 40	\\
                ARsmall      & 2600      	& 1260     	& 100      \\
			    LetterRecognition  & 780    	& 16     	& 26       \\
               Yale  & 165		& 256		& 15	\\
                Wine	  & 178		& 13		& 3	\\
                Feret   & 1400	  	& 1024	 	& 4	  \\
                Movement   &  360 & 90 & 15\\
                Ecoli & 336 & 7 &8\\
			    Yeast & 1484	  	& 1470      	& 10	  \\
			\bottomrule
	\end{tabular}}
\end{table}

\subsubsection{Compared Methods.}
To demonstrate the superiority of our algorithm ANCMM, we compare it with various state-of-the-art algorithms. We chose the K-Means algorithm, Spectral Clustering (SC), CAN \cite{nie2014clustering}, ERCAN  \cite{WANG2022108517} and DSDC \cite{Li2023TCSVT} for comparison. Among them, the CAN and ERCAN are adaptive neighbor algorithms. The DSDC algorithm is a  symmetric doubly stochastic algorithm.
When using each algorithm, we applied the same preprocessing to the data to ensure fairness.
\begin{table}[!htbp]
	\centering
	\caption{ ACC by our algorithm and comparison algorithms}
	\label{T2}
	\scalebox{0.8}{
		\begin{tabular}{@{}ccccccc@{}}
			\toprule
			Method     & K-Means & SC  & DSDC &CAN&ERCAN&OUR\\ 
			\midrule
			TR41&63.7&72.1&66.0&71.8&\textbf{75.1}&73.1\\
			ORL&56.3&53.5&60.5&58.0&57.0&\textbf{61.5}\\
			ARsmall&13.38&14.3&13.38&14.6&18.3&\textbf{15.81}\\
			LetterRecognition&32.7&33.9&34.7&32.1&31.0&\textbf{35.0}\\
	Yale&45.5&	47.9&	50.3&	\textbf{52.1}&	47.9&	\textbf{52.1}\\
			Wine&97.2&97.2&89.9&97.2&47.8&\textbf{98.3}\\
			Feret&29.6&25.9&29.7&29.7&25.6&\textbf{31.0}\\
			Movement&47.0&\textbf{50.8}&41.4&50.2&47.8&50.6\\
			Ecoli&80.4&56.6&54.2&83.6&57.4&\textbf{83.9}\\
			Yeast&48.7&42.0&34.0&48.9&43.3&\textbf{50.1}\\
			\bottomrule
	\end{tabular}}
\end{table}

\begin{figure}[h]
	\centering
	\subfigure[Ecoli]
	{
		\includegraphics[width=0.22\textwidth]{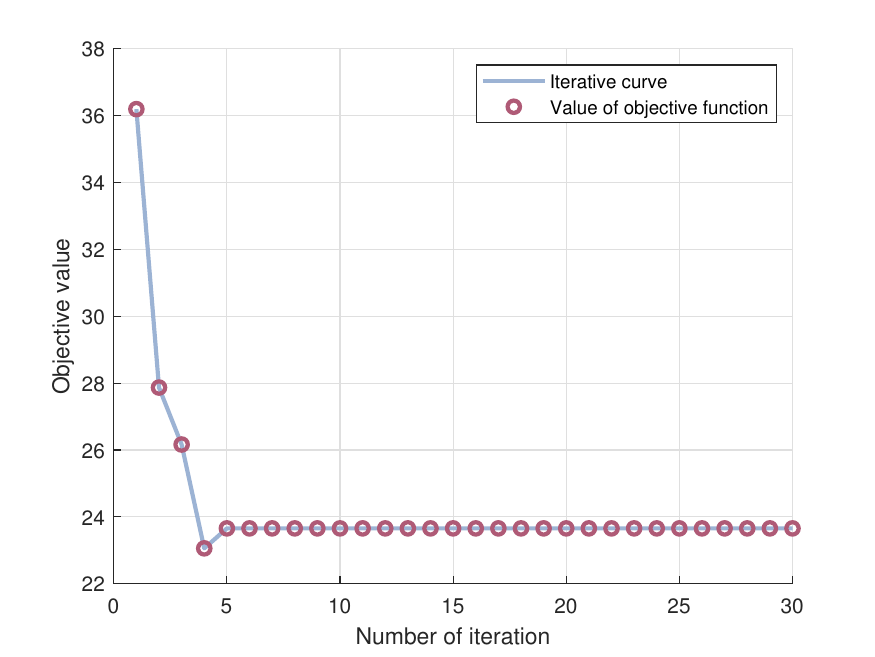}
	}
	\subfigure[LetterRecognition]
	{
		\includegraphics[width=0.22\textwidth]{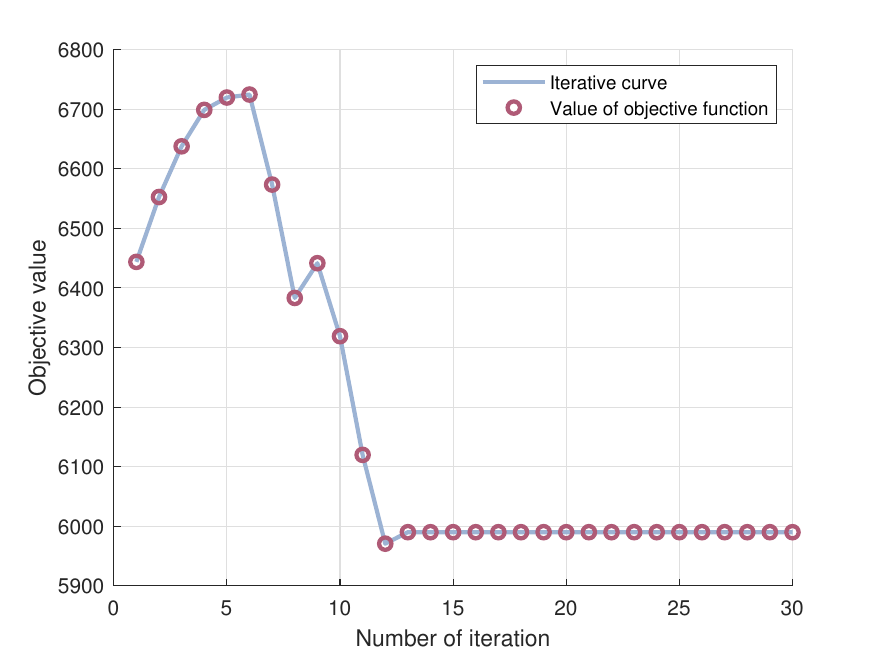}
	}
	\subfigure[Movement]
	{
		\includegraphics[width=0.22\textwidth]{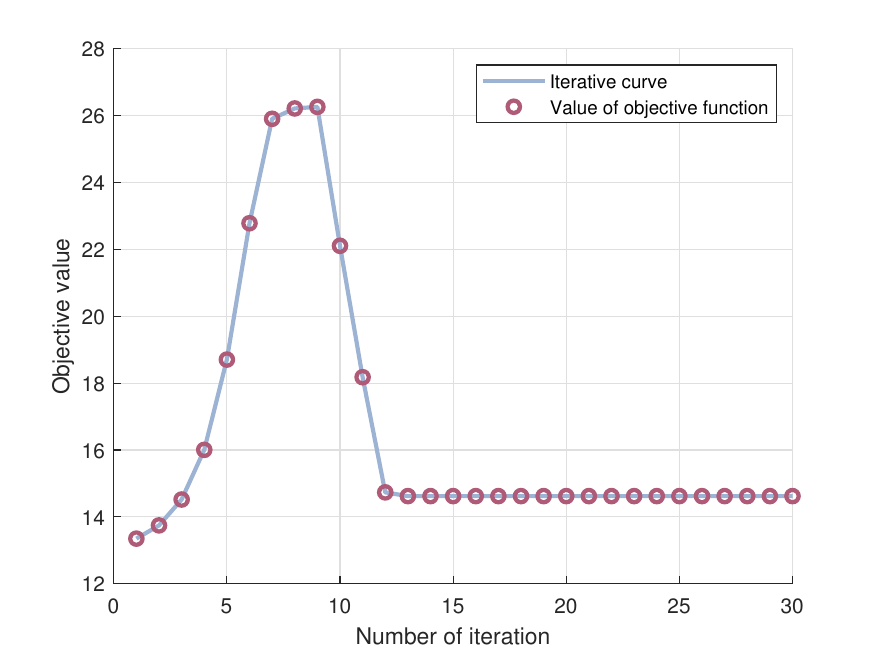}
	}
	\subfigure[Wine]
	{
		\includegraphics[width=0.22\textwidth]{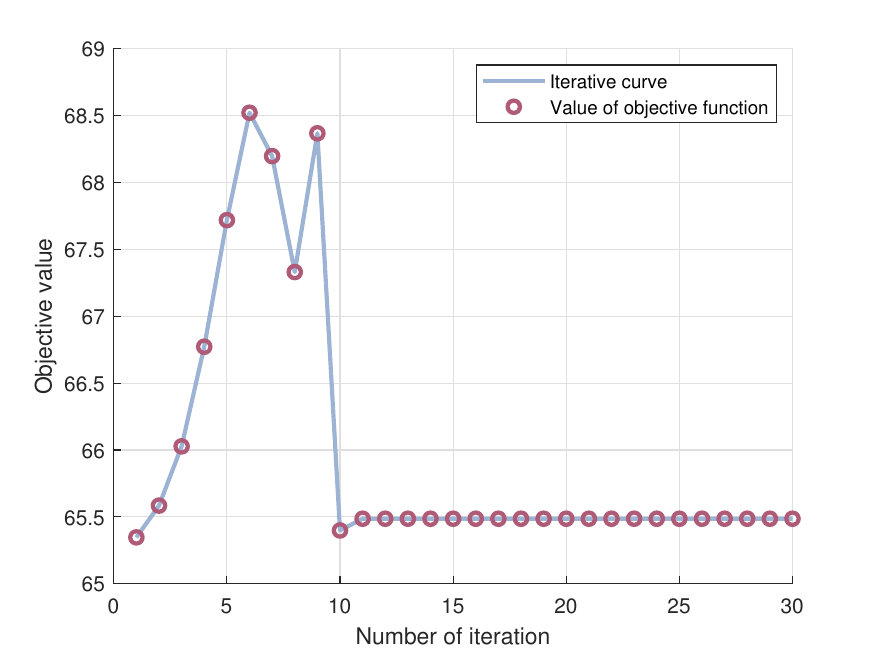}
	}
	\caption{Convergence experimental results on real-world datasets. (a) Ecoli. (b) LetterRecognition. (c) Movement. (d) Wine.}
	\label{F6}
\end{figure}

\subsubsection{Parameters Selection.}Our model has two parameters, $\alpha$ and $\lambda$. $\lambda$ is adaptive in the experiments and we randomly select a positive $\lambda$. If $rank(L_S)$ in each iteration is greater than $n-c$, we set $\lambda\leftarrow2\lambda$, otherwise, we set 
$\lambda\leftarrow\frac{1}{2}\lambda$, until final convergence. This provides a heuristic way to set \(\lambda\). For the other methods, we randomly initialize and take the best result after 30 runs.

\subsubsection{Evaluation Metrics.}In the experiments, we used accuracy (ACC), normalized mutual information (NMI), and purity (PUR) to evaluate the results. Tables 3, 4 and 5 present the corresponding results.
\begin{table}[!htbp]
	\centering
	\caption{ NMI by our algorithm and comparison algorithms}
	\label{T2}
	\scalebox{0.8}{
		\begin{tabular}{@{}ccccccc@{}}
			\toprule
			Method     & K-Means & SC  & DSDC &CAN&ERCAN&OUR\\ 
			\midrule
			TR41&67.3&70.9&67.4&72.3&\textbf{75.6}&74.7\\
			ORL&74.8&73.2&\textbf{76.7}&73.5&74.9&75.7\\
			ARsmall&42.3&\textbf{42.8}&40.9&37.0&37.3&38.6\\
			LetterRecognition&44.3&45.4&\textbf{46.7}&39.8&40.5&42.3\\
			Yale&52.9&	51.4&	53.9&	\textbf{60.5}&	54.3&	59.2\\
			Wine&89.0&89.0&73.2&89.0&83.0&\textbf{92.6}\\
			Feret&\textbf{67.2}&63.0&63.0&61.3&62.2&64.3\\
			Movement&59.0&60.9&52.8&63.5&62.4&\textbf{64.0}\\
			Ecoli&66.0&53.9&51.2&71.3&51.9&\textbf{73.6}\\
			Yeast&28.0&27.7&23.0&29.6&\textbf{47.9} &29.8\\
			\bottomrule
	\end{tabular}}
\end{table}
\begin{table}[!htbp]
	\centering
	\caption{ PUR by our algorithm and comparison algorithms}
	\label{T2}
	\scalebox{0.8}{
		\begin{tabular}{@{}ccccccc@{}}
			\toprule
			Method     & K-Means & SC  & DSDC &CAN&ERCAN&OUR\\ 
			\midrule
			TR41&83.5&85.3&85.1&84.7&85.8&\textbf{85.9}\\
			ORL&61.3&59.0&63.0&63.8&59.3&\textbf{64.8}\\
			ARsmall&13.8&14.6&13.9&16.0&19.1&\textbf{16.7}\\
			LetterRecognition&34.9&35.3&36.5&35.6&33.7&\textbf{38.9}\\
			Yale&48.5&	48.5&	51.5&	52.7&	50.3&	\textbf{52.7}\\
			Wine&97.2&97.2&89.9&97.2&94.9&\textbf{98.3}\\
			Feret&32.6&28.6&32.0&\textbf{34.6}&29.2&34.3\\
			Movement&50.0&51.7&46.1&\textbf{51.9}&51.1&51.7\\
			Ecoli&82.7&83.9&82.1&84.8&81.9&\textbf{85.7}\\
			Yeast&55.3&\textbf{55.3}&52.5&50.7&45.8&51.0\\
			\bottomrule
	\end{tabular}}
\end{table}
\subsection{Experimental Result}

To illustrate the convergence of our algorithm, we selected four datasets: Ecoli, Wine, Movement, and LetterRecognition. We plotted their convergence curves, as shown in Figure 2. According to the results, our algorithm typically converges in around 10 iterations.

\section{Conclusion}
This paper proves the theorems related to the Marcus mapping and proposes a clustering method based on Marcus mapping, called Doubly Stochastic Adaptive Neighbors Clustering (ANCMM). Under rank constraint conditions, the similarity matrix obtained can be directly partitioned into specific clusters, and the similarity matrix is doubly stochastic and symmetric. We also discuss the relationship between Marcus mapping and optimal transport, illustrating that Marcus mapping solves a special form of optimal transport, but the computation through Marcus mapping is superior to that through optimal transport. Future work can explore the application of Marcus mapping and rank constraints in other fields such as dimension reduction.
\subsection{Limitations}
Although the proposed method can learn symmetric doubly stochastic similarity graphs, several limitations remain. The optimization procedure relies on a row-wise relaxation followed by symmetrization and Marcus mapping, but no formal guarantee is provided that these operations solve the original coupled optimization problem or produce exactly the desired number of connected components. In addition, the regularization parameters and stopping criteria are selected heuristically.
\bibliography{aaai25.bib}

\end{document}